\documentclass[sigconf]{acmart}
\usepackage{algorithm}
\usepackage{algpseudocode}
\usepackage{amsmath}

\usepackage{amssymb}
\usepackage{multirow}
\usepackage{algpseudocode}
\usepackage{balance}
\AtBeginDocument{%
  }

\setcopyright{acmcopyright}
\copyrightyear{2018}
\acmYear{2018}
\acmDOI{XXXXXXX.XXXXXXX}

\acmConference[Conference acronym 'XX]{Make sure to enter the correct
  conference title from your rights confirmation emai}{June 03--05,
  2018}{Woodstock, NY}

\acmSubmissionID{594}

\copyrightyear{2023}
\acmYear{2023}
\setcopyright{acmlicensed}
\acmConference[MM '23] {Proceedings of the 31st ACM International Conference on Multimedia}{October 29--November 3, 2023}{Ottawa, ON, Canada.}
\acmBooktitle{Proceedings of the 31st ACM International Conference on Multimedia (MM '23), October 29--November 3, 2023, Ottawa, ON, Canada}
\acmPrice{15.00}
\acmISBN{979-8-4007-0108-5/23/10}
\acmDOI{10.1145/3581783.3611829}

\begin{document}

\title{Recurrent Spike-based Image Restoration under General Illumination}

\author{Lin Zhu}
\orcid{0000-0001-6487-0441}
\affiliation{%
  \institution{Beijing Institute of Technology}
  \city{Beijing}
  \country{China}
}
\email{linzhu@bit.edu.cn}

\author{Yunlong Zheng}
\orcid{0009-0008-1882-3124}
\affiliation{%
  \institution{Beijing Institute of Technology}
  \city{Beijing}
  \country{China}
}
\email{zhengyunlong@bit.edu.cn}

\author{Mengyue Geng}
\orcid{0000-0002-9480-2658}
\affiliation{%
  \institution{Peking University}
  \city{Beijing}
  \country{China}
}
\email{mygeng@pku.edu.cn}

\author{Lizhi Wang}
\orcid{0000-0002-1953-3339}
\affiliation{%
  \institution{Beijing Institute of Technology}
  \city{Beijing}
  \country{China}
 }
 \email{wanglizhi@bit.edu.cn}

\author{Hua Huang}
\orcid{0000-0003-2587-1702}
\affiliation{%
  \institution{Beijing Normal University}
  \city{Beijing}
  \country{China}
}
\email{huahuang@bnu.edu.cn}
\authornote{Corresponding author: Hua Huang.}

\renewcommand{\shortauthors}{Lin Zhu et al.}

\begin{abstract}
Spike camera is a new type of bio-inspired vision sensor that records light intensity in the form of a spike array with high temporal resolution (20,000 Hz).
This new paradigm of vision sensor offers significant advantages for many vision tasks such as high speed image reconstruction.
However, existing spike-based approaches typically assume that the scenes are with sufficient light intensity, which is usually unavailable in many real-world scenarios such as rainy days or dusk scenes.
To unlock more spike-based application scenarios, we propose a Recurrent Spike-based Image Restoration (RSIR) network, which is the first work towards restoring clear images from spike arrays under general illumination. 
Specifically, to accurately describe the noise distribution under different illuminations, we build a physical-based spike noise model according to the sampling process of the spike camera. Based on the noise model, we design our RSIR network which consists of an adaptive spike transformation module, a recurrent temporal feature fusion module, and a frequency-based spike denoising module. 
Our RSIR can process the spike array in a recursive manner to ensure that the spike temporal information is well utilized. 
In the training process, we generate the simulated spike data based on our noise model to train our network. 
Extensive experiments on real-world datasets with different illuminations demonstrate the effectiveness of the proposed network.
The code and dataset are released at \url{https://github.com/BIT-Vision/RSIR}.
\end{abstract}


\begin{teaserfigure}
\vspace{-4mm}
\begin{center}
  \includegraphics[width=0.73\textwidth]{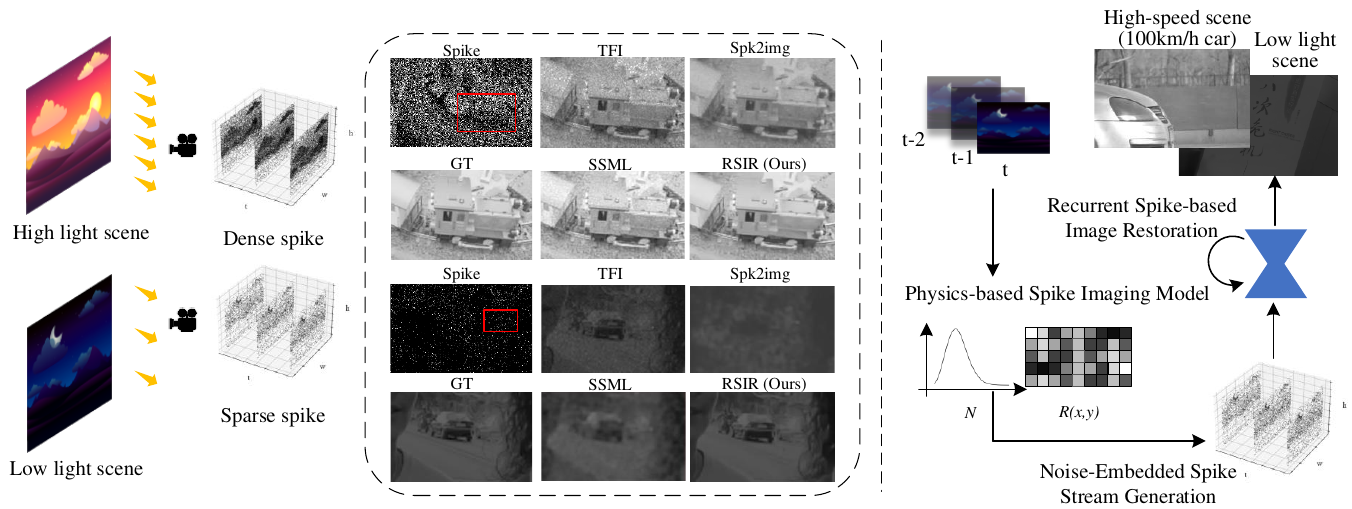}
\end{center}
\vspace{-4mm}
  \caption{Recurrent spike-based image restoration (RSIR) under general illumination. 
  (Left) Current spike-based restoration techniques (TFI~\cite{recon}, Spk2img~\cite{zhao2021spk2imgnet}, SSML~\cite{SSML}) struggle with low illumination conditions and are negatively affected by noise.
  (Right) Based on the proposed physics-based spike imaging model, we have developed a noise-embedded spike stream generation pipeline and RSIR model. Such realistic spike streams enable the effective training of our model, allowing it to restore clear images under a variety of lighting conditions.}
  \label{fig1}
\end{teaserfigure}


\maketitle

\section{Introduction}

In recent years, neuromorphic cameras have emerged as a groundbreaking innovation in the field of vision sensors, drawing inspiration from biological vision systems. These novel sensors offer significant advantages over traditional frame-based cameras, such as high dynamic range and high temporal resolution~\cite{Gehrig2018Asynchronous, Maqueda2018Event}. With diverse applications ranging from unmanned aerial vehicles and industrial monitoring to autonomous driving, neuromorphic cameras have shown immense potential~\cite{Litzenberger2006Embedded}. One crucial feature shared by all these cameras is their ability to asynchronously measure light intensity at the pixel level~\cite{Gehrig2018Asynchronous, Maqueda2018Event}.

Among various bio-inspired vision sensors, a notable development in this domain is the spike camera~\cite{recon}, which employs an integral sampling mechanism inspired by the fovea region of the retina. Spike cameras boast full texture sensing capability and high-speed sensing, allowing them to capture fast motion in the form of spike streams and paving the way for high-speed vision tasks.

Recent advancements include the development of a fovea-like texture reconstruction framework for image reconstruction~\cite{zhu2020retina,zhao2021spk2imgnet,recon}, as well as methods based on spike cameras for tone mapping~\cite{9181055} and motion deblurring~\cite{Han_2020_CVPR}. Nevertheless, most spike-based image reconstruction methods focus on scenes with sufficient light intensity. As spike cameras generate dense spikes in high illumination due to the integral sampling mechanism, they provide sufficient information for image reconstruction, with quantization noise being the primary source of noise distribution in such scenarios (as depicted in Fig.~\ref{fig:syn_comp}).

However, real-world scenes do not always have optimal lighting conditions. As illustrated in Fig.~\ref{fig1}, the spike densities differ considerably between high and low illumination levels. Existing methods face challenges in handling low illumination scenarios due to the stark differences in noise distributions. As a result, this paper tackles a novel question: \emph{Is it possible to restore high-quality images from spike streams under a wide range of lighting conditions, including both high and low light situations?}

In this paper, we propose a recurrent model to tackle the above challenge of spike-based image restoration.
To the best of our knowledge, this is the first work to address spike-based image restoration under general illumination.
Contributions of this paper are summarized as follows.

1) To address the typical challenge of illumination in spike-based reconstruction, we present a novel, recurrent approach for restoring visual images from continuous spike streams, which is specifically designed based on the noise in the sampling process of the camera. Therefore, our model can handle various illumination scenes, with a particular focus on scenes with insufficient illumination that are difficult to handle with existing methods.

2) We propose a novel physics-based spike noise model which is based on the real spike generation process. Our recurrent denoising framework consists of adaptive spike transformation, a temporal fusion module, and a spike denoising model, all of which are designed based on the proposed spike imaging model, which provides a clear objective for addressing different types of noise. 

3) We design a spike simulation pipeline based on the proposed noise model and construct a simulated spike dataset for training and testing our model. Additionally, we build a real spike dataset including various illumination levels. Experimental results from multiple real and simulated spike datasets demonstrate that our proposed model can effectively restore information from both dark and light areas from the spike stream.

\vspace{-1mm}
\section{Related Work}
\subsection{Noise Model in Traditional Cameras}

Image restoration based on fully supervised learning methods often involves an exceedingly complex and time-consuming process for obtaining paired real images~\cite{SID, SMID, SMOID, SIDD}. An alternative approach to forming paired training data is to model noise (either physically~\cite{ELD, RNS} or implicitly learned~\cite{NF, CANGAN, VDIR, DUS}) and synthesize noisy inputs similar to real captures. Simultaneously, the noise level derived from the noise model can be used as a parameter for adjusting denoising strength~\cite{CBDNet, FGELD, EMVD} or introducing an additional prior to constrain the network~\cite{VDN}.
Early image denoising work typically utilized a simple white Gaussian noise model to synthesize noise~\cite{BM3D, NLM, DnCNN}, which significantly differs from the distribution of true noise. Some works have focused on the Poisson-Gaussian model~\cite{PG} or a more realistic heteroscedastic Gaussian model~\cite{UNPROCESS, KPN}


\begin{figure*}[t]
\setlength{\abovecaptionskip}{0cm}
\setlength{\belowcaptionskip}{-0.2cm}
\begin{center}
\includegraphics[width=0.9\linewidth]{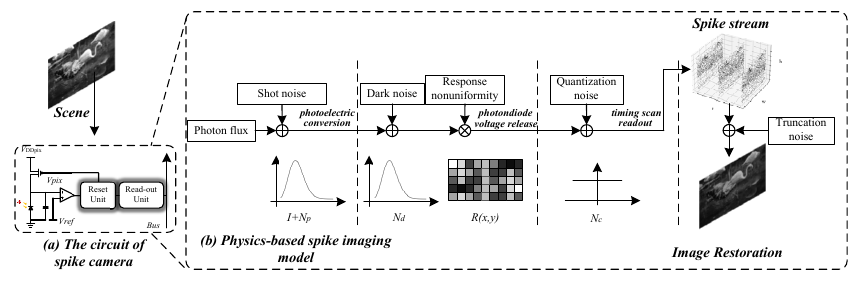}
\end{center}
\vspace{-3mm}
   \caption{Schematic diagram of spike camera's circuit and the physics-based spike imaging model.}
   \vspace{-3mm}
\label{fig:onecol}
\end{figure*}

\subsection{Image/Video Denoising}

Image denoising is an active task in low-level computer vision, with the widespread use of deep learning in low-level vision tasks, numerous methods based on different strategies have emerged.
More recently, a series of self-supervised methods based on untrained neural network prior~\cite{DIP, DIPRED, MCDIP, RDIP}, blind-spot networks~\cite{BSN, DBSN, APBSN}, masked pairs~\cite{Noise2Self, Noise2Void} or paired noisy-noisy data~\cite{Noisier2Noise, NoisyasClean} ave achieved relatively good results without using clean references.

The key to video denoising is leveraging the additional temporal information compared to single-image denoising. 
Motion estimation and compensation is an intuitive approach to promote beneficial information at the same pixel location between frames. 
This idea has been adopted by \cite{VDRME, VESPCN, DVD} based on optical flow, and \cite{EDVR, TDAN} based on deformable convolutional networks\cite{DCN}. 
However, these methods require more computational resources and may introduce additional errors from misalignment.
Many researchers have sought to handle video sequence more efficiently, using temporal and spatial attention mechanisms\cite{EDVR, PFNL}.
In addition, the extensibility of the recurrent structure makes it an important framework for video processing\cite{FRVSR, RTVD, EMVD}.

\subsection{Spike-based Image Restoration}
Image restoration based on spike camera has emerged in recent years. 
Texture images can be reconstructed by counting the spike interval and the number of spikes~\cite{recon}.
Based on the spiking neuron model, a fovea-like texture reconstruction framework was proposed to reconstruct images~\cite{zhu2020retina}. Some methods based on the spike camera were developed for tone mapping~\cite{9181055}, motion deblurring~\cite{Han_2020_CVPR}, super resolution~\cite{zhao2021super}.
Recently, deep learining based image reconstruction methods~\cite{zhao2021spk2imgnet,zhu2021neuspike,SSML} are proposed for spike camera, and their performance is much higher than that of traditional methods.
In fact, the noise type of spike cameras differs significantly from that of traditional cameras. A preliminary analysis of spike camera noise in~\cite{zhu2021neuspike} demonstrates that the noise distribution varies under bright and dim lighting conditions.
Given the lack of a dataset with ground truth images for spike cameras, investigating the noise distribution of spike cameras and generating realistic noisy spikes is crucial for spike-based applications.

In this work, we propose a noise model for spike cameras that considers the physical imaging process and circuit characteristics to synthesize noise distributions closely resembling real data.

\section{Physics-based Spike Imaging Model}
In this section, we revisit the working principle of the spike camera. Rather than focusing on biologically inspired modeling~\cite{recon, zhu2020retina}, we model the spike generation process from the perspective of physical imaging models. This will provide guidance for designing our network. Additional, we propose a realistic noise-embedded spike stream generation pipeline based on the spike noise model.

\subsection{Spike Signal Model}

Unlike the dynamic vision sensor (DVS), the spike camera reflects the light intensity of the scene by the discharge time when the photodiode voltage is released to the reference voltage.
Specifically, each pixel has three basic working processes (as shown in Fig.~\ref{fig:onecol}): integration, reset, readout.
In the integration phase, the photodiode accumulates electrons converted by photoelectric conversion, similar to conventional cameras. This process causes the voltage $V_{pix}$ at the input of the comparator to drop, and can be represented by the following equation:
\begin{equation}
    \setlength{\abovecaptionskip}{0pt}
    \setlength{\belowcaptionskip}{0pt}
    V_{pix}(t)=\int_{0}^{t}V_{DDpix}-\frac{I_{ph}(\tau)}{C}\,d\tau
\end{equation}
where $V_{DDpix}$ is the reset voltage, and $I_{ph}$ represents the magnitude of the current for photoelectric conversion.
$C$ is the capacitor in parallel with the photodiode, which is used to store electrons for photo-conversion.

When $V_{pix}$ drops to the reference voltage, the reset unit is triggered to perform a reset operation, sending a spike signal to the reset battery and the read-out unit.
The battery charges the capacitor, increasing $V_{pix}$ to $V_{DDpix}$, and the circuit enters in integration phase again.
The brightness of the scene is measured by the integration time $T$, which satisfies
\begin{equation}
    \int_{0}^{T}\frac{C\,\Delta V}{I_{ph}(\tau)}\,d\tau=1 \label{T_Idea}
\end{equation}
where $\Delta V=V_{DDpix}-V_{pix}$.

The flip-flop of the readout unit stores the spike signal sent by the reset unit and waits for the clock signal to output 0 or 1 to the bus.
We can use the step function $\epsilon(x)$ to simulate the potential changes, representing the process of obtaining the spike signal of the readout circuit followed by the clock signal.
The step function $\epsilon(x)$ is defined as:
\begin{equation}
    \epsilon(x)=\left\{
	\begin{aligned}
	1 \quad x\geq 0\\
	0 \quad x<0\\
	\end{aligned}
	\right.
\end{equation} 

Based on above analysis, the spike signal model can be defined as
\begin{equation}
    S(x,y,t)=S(x,y,nt_0)=\epsilon(\int_{t'}^{n t_0}\frac{C\,\Delta V}{I_{ph}(x,y,\tau)}\,d\tau-1)
\end{equation}
where $t'$ is the start time of the current integration phase, will be reset when the integration phase ends.
$n t_0$ indicates $n$-th the clock signal, and $t_0$ is the unit time interval of the clock signal, which equals $50\mu s$.

\subsection{Spike Noise Model}
Due to the significant differences in circuits, modeling the noise of a spike camera greatly differs from that of a traditional camera. We model spike cameras from a physical process perspective using a similar idea of modeling noise from traditional cameras.
\subsubsection{Shot Noise}\label{sec:ShotNoise}

Same as traditional cameras, the photosensor of spike camera receives photons from the scene, which are immediately converted into photo-electrons to form a photo-current in the circuit according to the photoelectric response.
Even in a uniform light scenario, the photons hitting the diode are not constant~\cite{CMOSISN}.
The difference between the number of photons at a given moment and the ideal number of photons is referred to as shot noise $N_{p}$:
\begin{equation}
    E+N_{p}\sim P(E)
\end{equation}
where $E$ is ideal number (or the mean number assuming the certain light intensity last infinity time) of photons hitting the photosensor. $P$ denotes the Poisson distribution.

\subsubsection{Dark Current Noise}\label{sec:DarkCurrentNoise}

Considering the thermal diffusion of charge carriers and defects inside or on the surface of PN junction, the spike circuit itself would output spike signals even if the scene light intensity is 0~\cite{CCDNOISE, CMOS}.
Given the randomness of generation and relevance with component, dark current noise $N_d$ can be modeled as (we ignore the effect of temperature):
\begin{equation}
    N_d(x,y)\sim P(\mu_d(x, y))
\end{equation}
where $\mu_d$ is the mean of $N_d$, and $(x,y)$ means $N_d$ and $\mu_d$ are pixel-specific.
$P$ denotes the Poisson distribution, representing dark current shot noise.

\begin{figure}[t]
\setlength{\abovecaptionskip}{0.2cm}
\setlength{\belowcaptionskip}{-0.5cm}
\begin{center}
\includegraphics[width=0.7\linewidth]{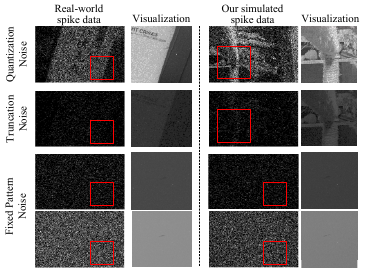}
\end{center}
\vspace{-3mm}
   \caption{The main types of noise in real and simulated spike data and visualization (\textbf{Best viewed with zoom in}).}
   \vspace{-1mm}
\label{fig:syn_comp}
\end{figure}

\subsubsection{Response Nonuniformity Noise}\label{sec:ResponseNonuniformityNoise}
Due to the limitations of the chip manufacturing process, the photodiode and capacitance mismatch will cause pixel units seem to be sensitive differently to light intensity, leading to photo response nonuniformity noise $N_{nru}$.
This mismatch would result in the rewriting of Eq. \ref{T_Idea} as:
\begin{equation}
    \int_{0}^{T+N_{rnu}}\frac{(C+\delta C)\,(\Delta V+\delta V)}{I_{ph}(\tau)}\,d\tau=1
\end{equation}
where $\delta C$, $\delta V$ represent the error of capacitance and voltage from mismatch, respectively.
To make the expression and parameter estimation feasible, we combine the two variables into one product variable, denoted as $R$.
\begin{equation}
    R(x,y) = \frac{(C+\delta C(x, y))\,(\Delta V+\delta V(x, y))}{C+\Delta V}
\end{equation}

To be clear, we cannot separately calibrate the specific errors of $\delta C$ and $\delta V$, so the existence of $R$ is necessary and not merely for the sake of formal brevity.

\subsubsection{Quantization Noise}\label{sec:QuantizationNoise}
Since the reset signal waits in the readout circuit for the clock signal to be output, the release time of the spike signal is always delayed relative to the generate time of the reset signal.
This process introduces quantization noise $N_q$:
\begin{equation}
    N_q \sim U(-t_0,t_0)
\end{equation}
where $U$ denotes the uniform distribution, and $t_0$ is clock signal interval.

The hysteresis of the readout circuit may ignore multiple reset signals in a clock time, but considering the clock time is small enough that such cases are almost non-existent under general illumination. This noise has a significant impact on strong light intensity.

\subsubsection{Truncation Noise from Spike to Image}
Previous sections (\ref{sec:ShotNoise}-\ref{sec:QuantizationNoise}) have analyzed the noise in the process of receiving photons and generating spike signals. 
However, the 1-bit high frame rate spike stream typically needs to be converted into a higher-bit image for additional post-processing.
Theoretically, using a longer spike stream for reconstruction would result in better visual quality in static scenarios. Nevertheless, in real-world scenarios, the impact of motion must be taken into account. 
For example, existing neural network based methods such as Spk2img only utilize information from a limited number of frames, often just tens of frames.
If the temporal length of the spike stream is not long enough, truncation noise will appear in the process from the spike to the image.

\begin{algorithm}[t]
\footnotesize
\caption{Realistic Noise-Embedded Spike Stream Generation Pipeline}\label{alg:syn}
\begin{algorithmic}[1]
\Require 
    A clean image $I$ that represents the light intensity of the scene;
    A light intensity factor $\theta$ used to simulate different illumination range;
    The length of simulated spike stream $l$;
\Ensure
    Simulated spike stream, $S$
\State Set the length of synthesized stream: $WINSIZE \gets l$
\State $I \gets I*\theta$
\While{$WINSIZE \neq 0$}
\State    Add shot noise: $I \gets  \mathcal{P}(I)$;
\State    Transform to charge time: $D \gets \frac{255}{I}$; 
\State    Add dark noise and nonuniformity noise: $D \gets \frac{D}{R+\frac{D\times \mathcal{P}(L_d)}{T_2\times (L_2+L_d)}}$; 
\State    Add quantization noise: $D \gets D + Random(-1, 1)$;
\State    Transform back to light intensity: $I \gets \frac{255}{D}$;
\State    Simulate the integration phase: $Integrator \gets Integrator + I$;
\State    Simulate the readout circuit: $S[n-WINSIZE] \gets Integrator \leq threshold$;
\State    Reset the integrator: $Integrator \gets Integrator - S[l-WINSIZE]*threshold$;
\State    $WINSIZE \gets WINSIZE-1$;
\EndWhile
\end{algorithmic}
\end{algorithm}


Assuming that under uniform light intensity, the time interval of the readout circuit is uniform $d$, and the truncation noise error introduced using the time interval window of $l$ ($kd<l\leq(k+1)d, k\in \mathbf{N}$) can be written as:
\begin{equation}
    \frac{1}{d} + N_c=
    \left \{
    \begin{aligned}
        \frac{k+1}{l}&,P=\frac{l-kd}{d}\\
        \frac{k}{l}\ \ \ &,P=\frac{(k+1)d-l}{d}\\
    \end{aligned}
    \right.
\end{equation}
where $P$ indicates the value probability.
This formula reveals that if the scene is dark (means longer $d$), the shorter the sample time window length, the lower the signal-to-noise ratio.

\begin{figure*}
    \setlength{\abovecaptionskip}{0.35cm}
    \setlength{\belowcaptionskip}{-0.2cm}
    \centering
    \includegraphics[width=0.9\linewidth]{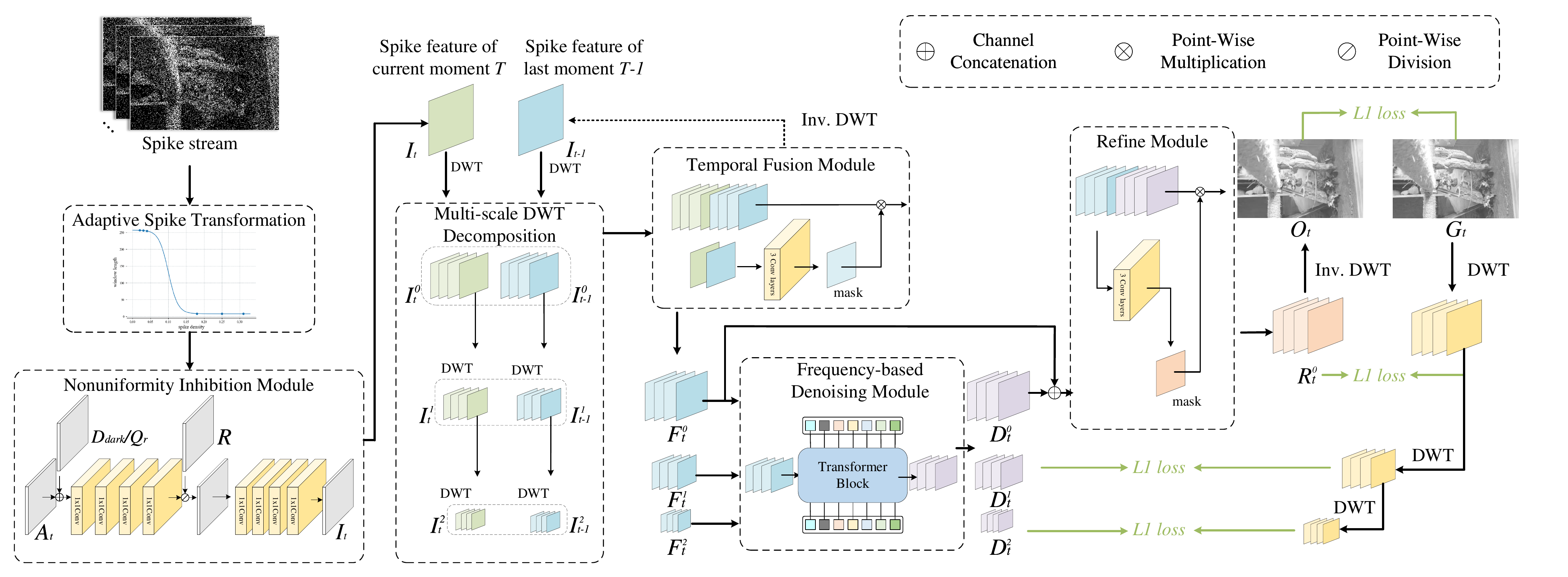}
    \vspace{-4mm}
    \caption{The structure of our recurrent spike-based image restoration model.}
    \vspace{-3mm}
    \label{fig:structure}
\end{figure*}

Based on above, our physics-based spike imaging model is defined as:
\begin{equation}
    I + N=\frac{1}{\frac{Q_r}{L+N_{p}+N_{d}}+N_{rnu}+N_{q}}+N_{c}
\end{equation}
where $I$ is the ideal image without noise, $N$ is the total noise, $L$ represents the scene light intensity, $Q_r$ is relative quantity matrix of electric charge.
$N_p$, $N_d$, $N_{rnu}$, $N_q$, and $N_c$ represent shot noise, dark current noise, response nonuniformity noise, quantization noise, and truncation noise, respectively.
\subsection{Noise Parameters Estimation}


Since spike cameras do not have additional analog gain or digital gain similar to conventional cameras, Poisson noise can be directly applied to a given light intensity (e.g., an image). 
This step introduces realistic photon shot noise.
 
The calibration of dark signals $\mu_d$ is also very different from that of conventional cameras.
Considering the variable discharge time $T$ that describes the brightness of the scene is inversely proportional to the electrons generated in the integration circuit, we cannot obtain the value of the dark signal just by only capturing the dark scene.
In addition to dark scenes, we also capture a uniform light scene and measured the photometric value using a photometer.
Then we can solve the equivalent light intensity value $L_d$ of the dark signal $\mu_d$ by reducing the following equations:
\begin{equation}
    \left\{
	\begin{aligned}
	C\,\Delta V&=\alpha L_d\,T_d\\
	C\,\Delta V&=\alpha (L_1+L_d)\,T_1\\
	\end{aligned}
	\right.
\end{equation}
where $T_d$, $T_1$ is mean discharge time when scene light intensity equals 0, $L_1$, and can be calculated by average interval time from the output spike stream.

The deviation matrix $R(x,y)$ corresponding to the response nonuniformity noise can be obtained by capturing a uniform light scene and recording the intensity. Choose the pixel ($x_m$, $y_m$) which is closest to the average response value as the reference pixel. $R(x,y)$ is then obtained by calculating the ratio of the reference pixel's response value to the response values of other pixels:
\begin{equation}
    R(x,y)=\frac{(L_2+L_d(x_m, y_m))T_2(x_m,y_m)}{(L_2+L_d(x, y))T_2}
\end{equation}

To gain a more intuitive understanding of different types of noise and validate the precision of our noise parameter estimation, three types of noise that exert the most significant influence on spike imaging are shown in Fig.~\ref{fig:syn_comp}. The fixed pattern noise, which comprises dark current noise and response nonuniformity noise, significantly impacts the integration process at low light levels. This limits the maximum dynamic range of spike cameras~\footnote{Please refer to our supplementary material for more details about noise parameter estimation and spike stream generation.}.

\subsection{Noise-Embedded Spike Stream Generation}\label{sec:syn}
Based on our physics-based spike imaging model, we propose a realistic noise-embedded spike stream generation pipeline, the specific synthesis steps are shown in Algorithm \ref{alg:syn}.

\begin{figure}
    \setlength{\abovecaptionskip}{0.2cm}
    \setlength{\belowcaptionskip}{-0.5cm}
    \centering
    \includegraphics[width=0.65\linewidth]{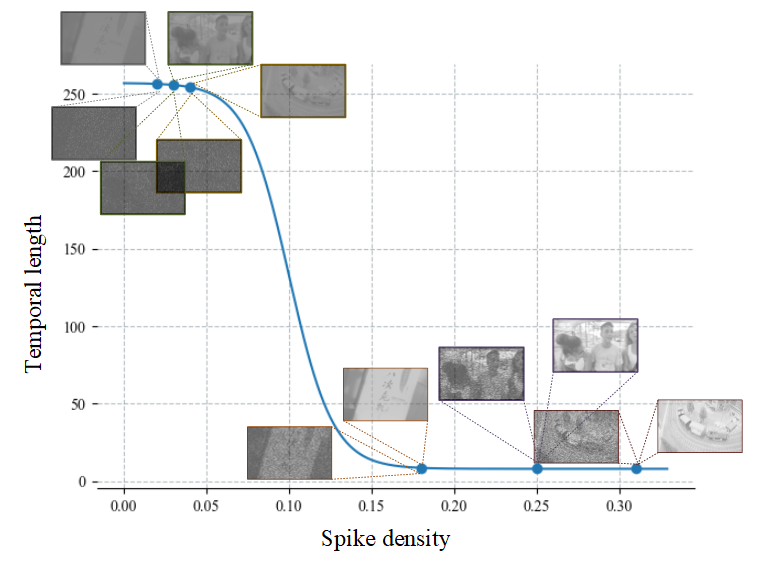}
    \vspace{-3mm}
    \caption{Adaptive spike transformation.}
    \vspace{-1mm}
    \label{fig:l-d}
\end{figure}
Each iteration of the loop represents a readout circuit process that verifies the reset signal in a clock signal. The image with an expanded range is interpreted as the scene's light intensity, and Poisson noise is added to it.
As dark current noise and response nonuniformity noise are evident in the discharge time, the light signal $I$ is transformed into a corresponding time interval signal $D$. The dark current noise, nonuniformity noise, and quantization noise are added to it before converting it back to the light signal.
Finally, by comparing the integrator's size with the threshold, it is determined whether a spike exists in the current clock signal. At the end of the loop, a noisy spike signal stream is generated.

\section{Recurrent Spike-based Image Restoration}
Based on the insights provided by our physics-based spike imaging model, we design our RSIR model. Our model is composed of various modules, each designed to effectively handle different types of noise.
\subsection{Adaptive Spike Transformation}\label{sec:ast}

To reduce truncation noise while providing sufficient light information, we propose adaptive spike transformation (AST), the window length $l$ - light intensity $L$ function.
AST select adaptive window by evaluating the relationship between light intensity and spike density.
The implementation of pixel-wise light intensity to dynamically adjust the window length can help our model better learning the spike feature:
\begin{equation}
    l(x,y)=AST(L(x,y))
\end{equation}

Pixel-specific windows allow more time domain noise removal without introducing blur throughout the scene.
In the specific selection of the form of AST, we mainly consider two kinds of prior: one is under the illuminance condition of more genral, dark light, spike camera has a very small response, resulting a high level of cut noise and extremely low and signal-to-noise ratio, which means we need a longer window to remove time domain noise;
Second, in the use of actual spike cameras, the moving scene often requires higher illumination conditions to capture sufficient light intensity information at a smaller motion amplitude.
At the same time, the signal-to-noise ratio is high and has less cut noise, so high light intensity requires a shorter window length
Based on the above, we propose a function much like Sigmoid to select the window length:

\begin{equation}
    AST(L) = -249*\frac{1}{1+\mathrm{e}^{-75*L+7.5}}+257
\end{equation}

The result is need to be rounded down, cause the window length is always an integer.

We calculate the mean of input spike stream in the selected window $l$ to get $A_t$ for the next processing of the network:
\begin{equation}
    A_t=mean(S[t-\frac{l}{2}:t+\frac{l}{2}])
\end{equation}

\begin{figure*}
    \setlength{\abovecaptionskip}{0.1cm}
    \setlength{\belowcaptionskip}{-0.2cm}
    \centering
    \includegraphics[width=0.805\linewidth]{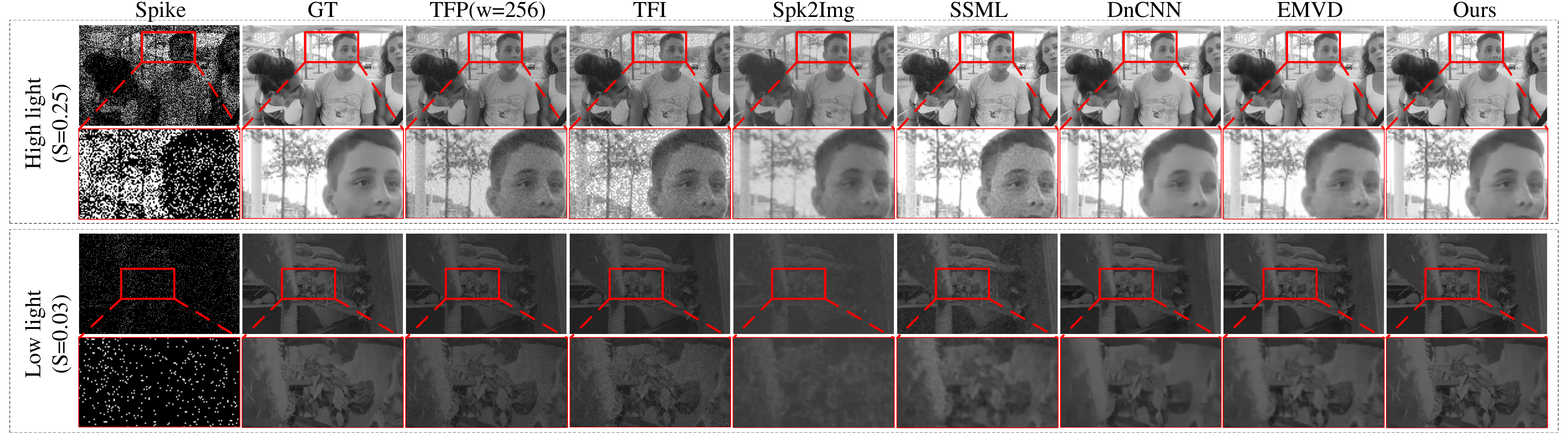}
    \vspace{-1mm}
    \caption{Comparison on simulated dataset (\textbf{Best viewed with zoom in}).}
    \vspace{-1mm}
    \label{fig:expsim}
\end{figure*}

\begin{figure*}
    \setlength{\abovecaptionskip}{0.1cm}
    \setlength{\belowcaptionskip}{-0.2cm}
    \centering
    \includegraphics[width=0.8\linewidth]{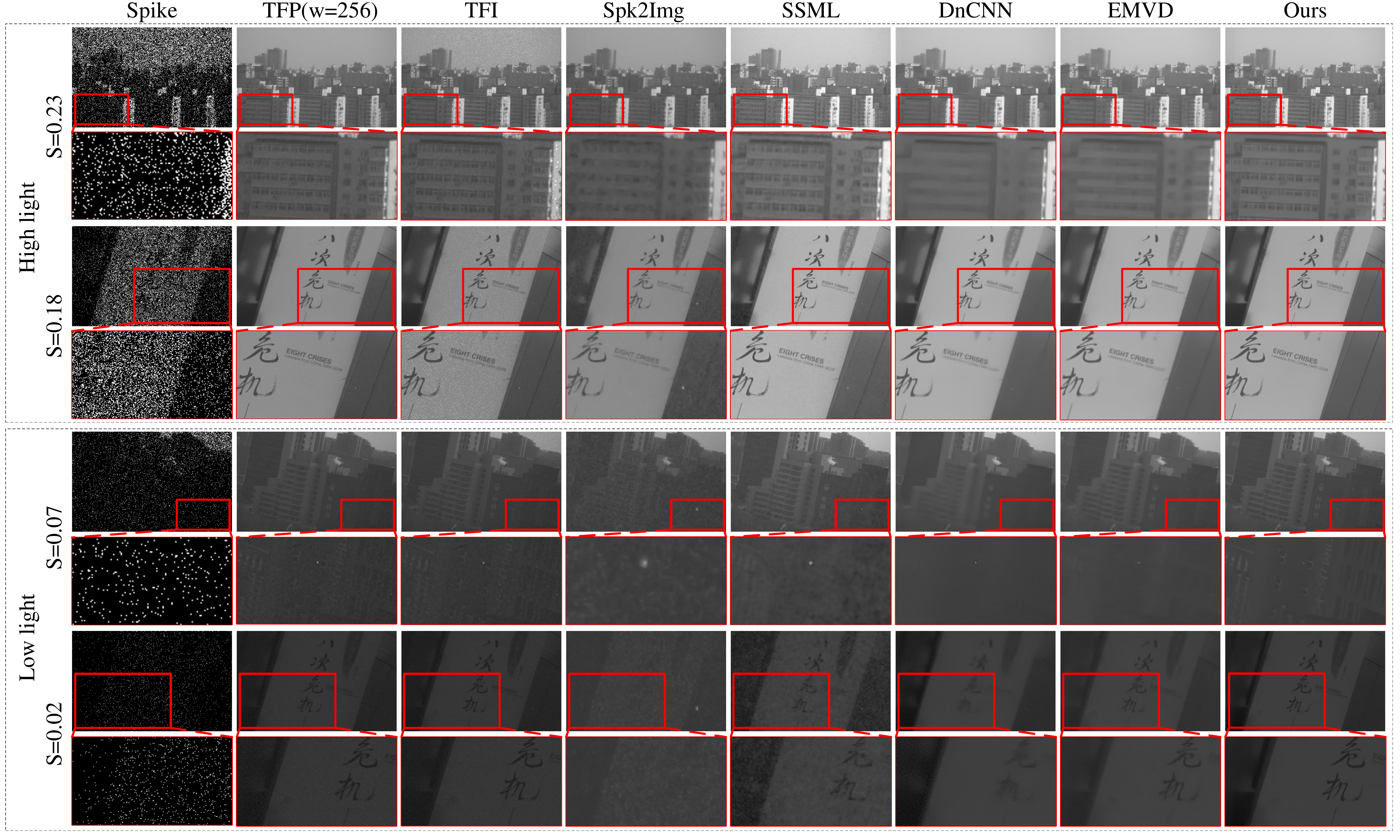}
    \vspace{-1mm}
    \caption{Comparison on real-world dataset (\textbf{Best viewed with zoom in}).}
    \label{fig:expreal}
\end{figure*}

The AST function and several real and synthetic scenes with different spike densities as shown in the Fig. \ref{fig:l-d}.
It can be seen that our synthetic spike have similar densities to real, and high-brightness and low-brightness spike have similar noise intensities after reconstruction with selected windows.
$A_t$ is fed to the NIM module mentioned in Sec. \ref{sec:SDM} to remove fixed pattern noise, resulting in $I_t$.

\subsection{Recurrent Denoising Framework}

Inspired by traditional video denoising, we adopt a recurrent denoising framework (RDF) like video processing to achieve spike denoising in a temporal recurrent manner.
As shown in Fig. \ref{fig:structure}, our model receives the image $I_t$ after adaptive spike transformation, as well as the $F_{t-1}$ from the last inference.
So the total input to our recurrent framework can be defined as:
\begin{equation}
    I = Concat(I_t,I_{t-1}),\ \ I_{t-1}=\left \{ 
    \begin{aligned}
        I_t\ \ &,t=0\\
        F_{t-1}&,otherwise\\
    \end{aligned}
    \right.
\end{equation}

To iterate the model at initial time, $I_t$ is used to instead of the non-existent $F_{t-1}$ when $t=0$.

To guide restoration with multiple scales and perform noise separation, we use discrete wavelet transform for downsampling the spike feature:
\begin{equation}
    {{I_t}^0}_{\{LL,LH,HL,HH\}}=\mathcal{DWT}(I_t)
\end{equation}
\begin{equation}
    {{I_t}^1}_{\{LL,LH,HL,HH\}}=\mathcal{DWT}({{I_t}^0}_{LL})
\end{equation}
\begin{equation}
    {{I_t}^2}_{\{LL,LH,HL,HH\}}=\mathcal{DWT}({{I_t}^1}_{LL})
\end{equation}

where ${I_t}^n$ denotes $n$ downsampling level of spike feature at time $t$, $LL$, $LH$, $HL$, and $HH$ denote the components of DWT.  $I_{t-1}$ has also undergone the above transformation.

\subsection{Temporal Fusion Module}

In order to remove the time-domain noise such as shot noise and truncation noise, we use the a fusion convolutional neural network (FCNN) inspired by \cite{EMVD} in the frequency domain.
We use a simple network with three convolutional layers to estimate the fusion $mask$.
The original frequency domain features would multiply $mask$ to obtain the output of the fusion module, this process can be expressed as:
\begin{equation}
    mask^2=\mathcal{FCNN}({I_{t}^2}_{LL}, {I_{t-1}^2}_{LL})
\end{equation}
\begin{equation}
    F_t^2=mask*I_t^2+(1-mask)*I_{t-1}^2
\end{equation}

The above illustrates the bottom-scale (2-$th$ scale) fusion process, and the higher scales will input an additional feature, the inverse DWT result from the lower fusion module output.

\subsection{Frequency-based Spike Denoising Module}\label{sec:SDM}

To suppress the effect of fixed pattern noise (dark current noise and response nonuniformity noise shown in Fig.~\ref{fig:syn_comp}), we propose a nonuniformity inhibition module (NIM)  containing several convolution layers.
As shown in Fig. \ref{fig:structure}, this module first concatenate  $I_t$ and $D_{dark}/Q_r$ together to fed into a series of convolution layers.
$D_{dark}$ is the time interval corresponding to the dark current, which can be calculated by capturing two different uniform light scenes.
$Q_r$ is the relative quantity matrix of electric charge, which describes the relative amount of charge of the circuit corresponding to each pixel in the sensor array in one discharge~\footnote{Please refer to our supplementary material for more details about $Q_r$.}.

The middle feature map obtained by dividing with $R(x,y)$ is fed into another several layers of convolution to suppress nonuniformity noise.

However, since the NIM module does not completely separate the truncation noise and quantization noise, there is still a small amount of fixed pattern noise and temporal noise after fusion stage.
We use Swin-transformer blocks~\cite{Swin-T} in the frequency domain to achieve more thorough noise removal.
At the scale 2, the process of the denoise module can be expressed as:
\begin{equation}
    {D_t}^2=\mathcal{SWIN-T}({F_t}^2)
\end{equation}

For seeking denoising prior guidance, the inputs of higher scales contain the output of the lower scale denoising module addtionally.

To eliminate additional artifacts and loss of image details from denoise, we use a simple CNN with residual connection to refine scale 0 fusion and denoise results.
\begin{equation}
    mask = \mathcal{RCNN}({F_t}^0,{D_t}^0)
\end{equation}
\begin{equation}
    {R_t}^0 = mask*{F_t}^0+(1-mask)*{D_t}^0
\end{equation}
where the structure of $\mathcal{RCNN}$ is exactly the same as $\mathcal{FCNN}$ except for the different input dimensions.
To obtain the final reconstructed picture, $R_t$ go through the inverse DWT operator:
\begin{equation}
    O_t = Inv.\,\mathcal{DWT}({R_t}^0)
\end{equation}

Finally, we use $L1$ loss to supervise the restored image and ground truth, and their DWT components on three scales as shown in Fig.~\ref{fig:structure}.

\section{Experiment}

\subsection{Experiment Setup}
We compare the proposed RSIR against various spike restoration methods and image denoise methods, including TFP~\cite{recon} with different window length, TFI~\cite{recon}, Spk2Img~\cite{zhao2021spk2imgnet}, SSML~\cite{SSML}, DnCNN~\cite{DnCNN} and EMVD~\cite{EMVD}.
In our experiments, we show the performance of each method on synthesised data and real data, considering the performance differences of high/low illumination scenes at meantime.
In order to ensure the advantages of spike cameras in scenes of high speed movement, we also verify the compatibility of our method for high speed motion scenes.
Finally, we conduct ablation study to find out the effectiveness of different modules and recurrent framework.

\begin{table}[t]
\setlength{\abovecaptionskip}{0cm}
\setlength{\belowcaptionskip}{-0.2cm}
\footnotesize
\centering
\caption{Quantitative evaluation on simulated dataset.}
\label{table1}
\begin{tabular}{c|c|c|c|c|c|c}
\hline
\multirow{3}*{Method} & \multicolumn{6}{c}{Metrics}\\
\cline{2-7}
& \multicolumn{2}{c|}{Low light} & \multicolumn{2}{c}{High light} & \multicolumn{2}{|c}{Mean}\\
\cline{2-7}
& PSNR & SSIM  & PSNR & SSIM & PSNR & SSIM \\
\hline
TFP(w=32) & 22.26 & 0.4264 & 18.45 & 0.7543 & 20.36 & 0.5904 \\
TFP(w=64) & 26.08 & 0.6965 & 18.58 & 0.7867 & 22.23 & 0.7416\\
TFP(w=128) & 27.75 & 08464 & 18.61 & 0.7987 & 23.18 & 0.8225\\
TFP(w=256) & 28.06 & 0.8998 & 18.62 & 0.8023 & 23.34 & 0.8510\\
TFI & 27.68 & 0.8347 & 17.75 & 0.6228 & 22.72 & 0.7287 \\
Spk2Img & 25.88 & 0.7176 & 18.78 & 0.8238 & 22.27 & 0.7714 \\ 
SSML & 31.77 & 0.8392 & 28.01 & 0.8375 & 29.89 & 0.8383\\ 
DnCNN & 34.84 & 0.8869  &32.43 & 0.9061  & 33.63 & 0.8965 \\ 
EMVD & 38.51 & 0.9451  & 33.36 & 0.9201  & 35.93 & 0.9326 \\ \hline
Ours & \textbf{41.94} & \textbf{0.9739} & \textbf{35.12} & \textbf{0.9410} & \textbf{38.53} & \textbf{0.9575} \\
\hline
\end{tabular}
\label{table:comparison of strategies}
\end{table}

Our implementation is based on PyTorch, with $L_1$ loss and Adam optimizer~\cite{kingma2014adam} used in training.
We run our RSIR on 1 NVIDIA GeForce RTX 3090 GPU with batch size 8, learning rate $10^{-4}$ without decay.

\subsection{Simulated Spike Dataset}

On the simulated data, we test the performance of all comparison methods, and metrics results of PSNR, SSIM are shown in the Table \ref{table1}.
The metrics value of ours are much higher than all other methods in both high-light and low-light scenes.
TFI and TFP methods only use the basic correspondence between spike and light intensity to construct image, so their performance is not good.
Spk2Img and SSML consider very limited scene applications, and they are only useful in high speed motion scenes under high illuminations.
When scene noise is taken into account under more general illuminations, they fail.
DnCNN and EMVD are classic algorithms in the field of image denoising.
We retrain them on the TFP restoration results from simulated dataset and compare them to verify that our model perform well not just only because we consider the noise under general illumination, but also because our model itself has a stronger fitting ability for noise-clean mapping.

The visual results on the simulated data is shown in the Fig. \ref{fig:expsim}, here shows the results of a bright scene (spike density equals 0.25) and a dark scene (spike density equals 0.03).
The traditional methods TFP, TFI has no loss in structural details, but they can not remove fixed pattern noise, TFI can not remove the quantization noise of the bright scene at meantime.
Spk2Img and SSML completely fail on the dark scene, even have much more noise than TFI for bright test scene.
DnCNN and EMVD try to remove fixed pattern noise with image local information, but their results are blurry and over-smoothed due to their simple structure.
Our results are very close to truth images with slight loss of details.

\begin{table}[t]
\setlength{\abovecaptionskip}{0cm}
\footnotesize
\centering
\caption{Quantitative evaluation on simulated and real dataset.}
\label{table1}
\begin{tabular}{c|c|c|c}
\hline
\multirow{2}*{Method} & \multicolumn{3}{c}{Brisque
metric $\downarrow$}\\
\cline{2-4}
& \multicolumn{1}{c|}{Simulated dataset} & \multicolumn{1}{c}{Real-world dataset} & \multicolumn{1}{|c}{Mean}\\ \hline
TFP(w=32) &38.99 & 43.04 & 41.01  \\
TFP(w=64) & 38.63 & 39.03 & 38.83 \\
TFP(w=128) & 33.56 & \underline{29.59} & 31.58 \\
TFP(w=256) & 29.69 & \textbf{26.73} & \textbf{28.21} \\
TFI & 36.72 & 34.81 & 35.77  \\
Spk2Img & 31.25 & 36.89 & 34.07\\ 
SSML & 30.04 & 30.06 & 30.05\\ 
DnCNN & 27.04 & 34.89  & 30.97 \\ 
EMVD & \textbf{22.73} & 42.35  & 32.54  \\ \hline
Ours & \underline{26.12} & 33.33 & \underline{29.72}  \\
\hline
\end{tabular}
\label{table:Brisque}
\vspace{-1em} 
\end{table}

\begin{table}[t]
\setlength{\abovecaptionskip}{0cm}
\footnotesize
\centering
\caption{Quantitative evaluation on high speed scenes.}
\label{table1}
\begin{tabular}{c|c|c|c|c|c}
\hline
Method & TFP & TFI & Spk2Img & SSML & Ours \\ \hline
Brisque & 20.48 & 40.21 & 35.91 & 31.35 & \textbf{19.21}\\
\hline
\end{tabular}
\label{table:expmotion}
\vspace{-1em} 
\end{table}


\subsection{Real-world Spike Dataset}

In order to verify that the restoration of our model on real scenes is also effective, and also to illustrate the rationality of our noise model and the synthetic noise data is realistic, we tested all comparison methods on real scenes taken by ourselves.

The visual results are shown in Fig. \ref{fig:expreal}.
We used Brisque metric to assist the comparison, and the results are shown in the Table \ref{table:Brisque}.
Since the image quality analyze without reference is still a challenge, visual comparison is more illustrative of the differences between the methods.
The results of each method in real scenarios are similar to those on simulated data basiclly, which once again proves that the noise model and synthesis method proposed by us are effective.
Our results can effectively remove fixed pattern noise and temporal noise while retaining most details. However, in extremely dark areas, some details may be smoothed out. Considering the difficulty of recovering details in such dark areas, we believe this is an acceptable trade-off.

\subsection{Comparison on High Speed Scenes}
\begin{figure}
    \setlength{\abovecaptionskip}{0cm}
    \setlength{\belowcaptionskip}{-0.2cm}
    \centering
    \includegraphics[width=1\linewidth]{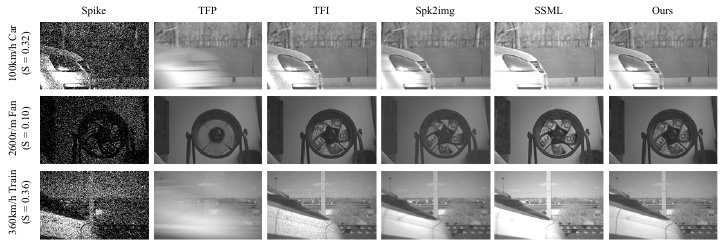}
    \caption{Comparison on real high speed scenes.}
    \label{fig:expmotion}
    \vspace{-1em} 
\end{figure}
The advantage of a spike camera lies in high-speed motion scenes. To validate that our network model can also handle high-speed motion, we compared our method with a pulse-image reconstruction method on existing real motion scenes.
The visual results are shown in Fig. \ref{fig:expmotion}, and Brisque metric values are shown in Table \ref{table:expmotion} 
The results show that our method can still remove the vast majority of temporal noise.

\subsection{Ablation Study}

\noindent\textbf{Effectiveness of the Proposed Denoising Pipeline}
Since our method is based on the frequency domain, all the intermediate features can be visualized by inverse DWT, and the metrics values of them to the truth image are calculated as shown in the Table \ref{table:comparison of different modules}.

\begin{table}[t]
\setlength{\abovecaptionskip}{0cm}
\footnotesize
\centering
\caption{Quantitative evaluation on simulated dataset.}
\label{table1}
\begin{tabular}{c|c|c|c|c|c}
\hline
Module & input & NIM & Fusion & Denoise & Refine\\
\hline
PSNR&20.07&23.31&31.32&38.50&{38.53}\\

SSIM&0.5344&0.5258&0.8480&0.9572&{0.9575}\\ \hline
\end{tabular}
\label{table:comparison of different modules}
\vspace{-2em}
\end{table}

\begin{figure}
    \setlength{\abovecaptionskip}{0cm}
    \setlength{\belowcaptionskip}{-0.2cm}
    \centering
    \includegraphics[width=\linewidth]{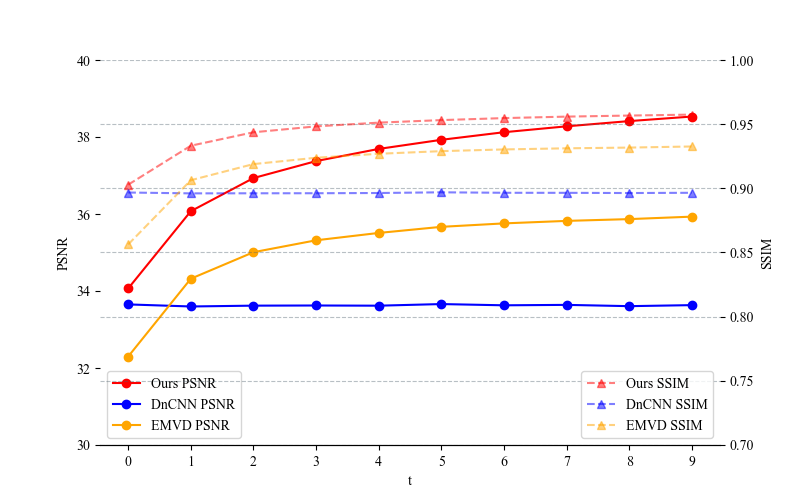}
    \caption{Comparison of different methods in temporal domain.}
    \label{fig:recurrent}
    \vspace{-1em} 
\end{figure}

\noindent\textbf{Effectiveness of Recurrent Mechanism}
As shown in Fig. \ref{fig:recurrent}, different methods process the results of spike stream from different time.
Our method can obtain higher metrics values over time, which shows that our recurrent structure is effective.

\noindent\textbf{Effectiveness of Adaptive Spike Transformation}
We propose the adaptive spike transformation to guarantee approximate noise levels between low-light and high-light inputs.
When we use the same window length to process high light and low light spike stream, the final result get worse.
Table \ref{table:comparison of strategies} shows the results of whether we use AST or not.


\begin{table}[t]
\setlength{\abovecaptionskip}{0cm}
\centering
\caption{Effectiveness of different strategies.}
\label{table1}
\begin{tabular}{c|c|c}
\hline
\multirow{2}{*}{Strategies}&\multicolumn{2}{|c}{Metrics}\\
\cline{2-3}
& PSNR & SSIM\\
\hline
w/o AST & 35.58 & 0.9367\\
w/ AST & \textbf{38.53} & \textbf{0.9575}\\
\hline

\end{tabular}
\label{table:comparison of strategies}
\vspace{-2em} 
\end{table}

\section{Conclusion}
We have presented a novel spike-based recurrent neural network
for image restoration under general illumination. 
By introducing the physics-based spike imaging model, we have developed a noise-embedded spike stream generation pipeline.
To take the advantage of the guidance of spike imaging model, we propose an RSIR model which consists of an adaptive spike transformation module, a recurrent temporal feature fusion module, and a frequency-based spike denoising module.
All of the proposed modules are designed based on the proposed spike imaging model, which provides a clear objective for addressing different types of noise.
We further show that the proposed framework generalizes well to multiple illuminations, while performing significantly better than recent state-of-the-art methods. 

\begin{acks}
    This work is supported by National Natural Science Foundation of China (62131003, 62072038) and Beijing Institute of Technology Research Fund Program for Young Scholars.
\end{acks}

\bibliographystyle{ACM-Reference-Format}
\balance
\bibliography{acmart}
\end{document}